\documentclass[sigconf]{acmart}

\usepackage{booktabs} 
\usepackage{graphicx, epstopdf}
\usepackage{subfigure}
\usepackage{url}
\usepackage{multirow}
\usepackage{algorithm}
\usepackage{algorithmic} 
\usepackage{balance}
\usepackage{arydshln}

\begin{document}
\title{Deep Spatio-Temporal Neural Networks for Click-Through Rate Prediction}
\author{Wentao Ouyang, Xiuwu Zhang, Li Li, Heng Zou, Xin Xing, Zhaojie Liu, Yanlong Du}
\affiliation{%
  \institution{Intelligent Marketing Platform, Alibaba Group}
  \city{Beijing, China}
}
\email{{maiwei.oywt, xiuwu.zxw, ll98745, zouheng.zh, xingxin.xx, zhaojie.lzj, yanlong.dyl}@alibaba-inc.com}

\begin{abstract}
Click-through rate (CTR) prediction is a critical task in online advertising systems. A large body of research considers each ad independently, but ignores its relationship to other ads that may impact the CTR. In this paper, we investigate various types of auxiliary ads for improving the CTR prediction of the target ad. In particular, we explore auxiliary ads from two viewpoints: one is from the spatial domain, where we consider the \emph{contextual} ads shown above the target ad on the same page; the other is from the temporal domain, where we consider historically \emph{clicked} and \emph{unclicked} ads of the user. The intuitions are that ads shown together may influence each other, clicked ads reflect a user's preferences, and unclicked ads may indicate what a user dislikes to certain extent. In order to effectively utilize these auxiliary data, we propose the Deep Spatio-Temporal neural Networks (DSTNs) for CTR prediction. Our model is able to learn the interactions between each type of auxiliary data and the target ad, to emphasize more important hidden information, and to fuse heterogeneous data in a unified framework. Offline experiments on one public dataset and two industrial datasets show that DSTNs outperform several state-of-the-art methods for CTR prediction. We have deployed the best-performing DSTN in Shenma Search, which is the second largest search engine in China. The A/B test results show that the online CTR is also significantly improved compared to our last serving model.
\end{abstract}



\copyrightyear{2019}
\acmYear{2019}
\setcopyright{acmcopyright}
\acmConference[KDD '19]{The 25th ACM SIGKDD Conference on Knowledge Discovery and Data Mining}{August 4--8, 2019}{Anchorage, AK, USA}
\acmBooktitle{The 25th ACM SIGKDD Conference on Knowledge Discovery and Data Mining (KDD '19), August 4--8, 2019, Anchorage, AK, USA}
\acmPrice{15.00}
\acmDOI{10.1145/3292500.3330655}
\acmISBN{978-1-4503-6201-6/19/08}

\settopmatter{printacmref=true}
\fancyhead{}

\maketitle

\section{Introduction}
Click-through rate (CTR) prediction is to predict the probability that a user will click on an item. It plays an important role in online advertising systems. For example, the ad ranking strategy generally depends on CTR $\times$ bid, where bid is the benefit the system receives if an ad is clicked. Moreover, according to the common cost-per-click charging model, advertisers are only charged once their ads are clicked by users. Therefore, in order to maximize the revenue and to maintain a desirable user experience, it is crucial to estimate the CTR of ads accurately.

CTR prediction has attracted lots of attention from both academia and industry \cite{he2014practical,cheng2016wide,shan2016deep,he2017neural,zhou2018deep}. One line of research is to take advantage of machine learning approaches to predict the CTR for each ad independently. For example, Factorization Machines (FMs) \cite{rendle2010factorization} are proposed to model pairwise feature interactions in terms of the latent vectors corresponding to the involved features.
In recent years, Deep Neural Networks (DNNs) are exploited for CTR prediction and item recommendation in order to automatically learn feature representations and high-order feature interactions \cite{van2013deep,zhang2016deep,covington2016deep}. To take advantage of both shallow and deep models, hybrid models are also proposed. For example, Wide\&Deep \cite{cheng2016wide} combines Logistic Regression (LR) with DNN, in order to improve both the memorization and generalization abilities of the model.
DeepFM \cite{guo2017deepfm} combines FM with DNN, which further improves the model ability of learning feature interactions.

\begin{figure}[!t]
\centering
\includegraphics[width=0.48\textwidth]{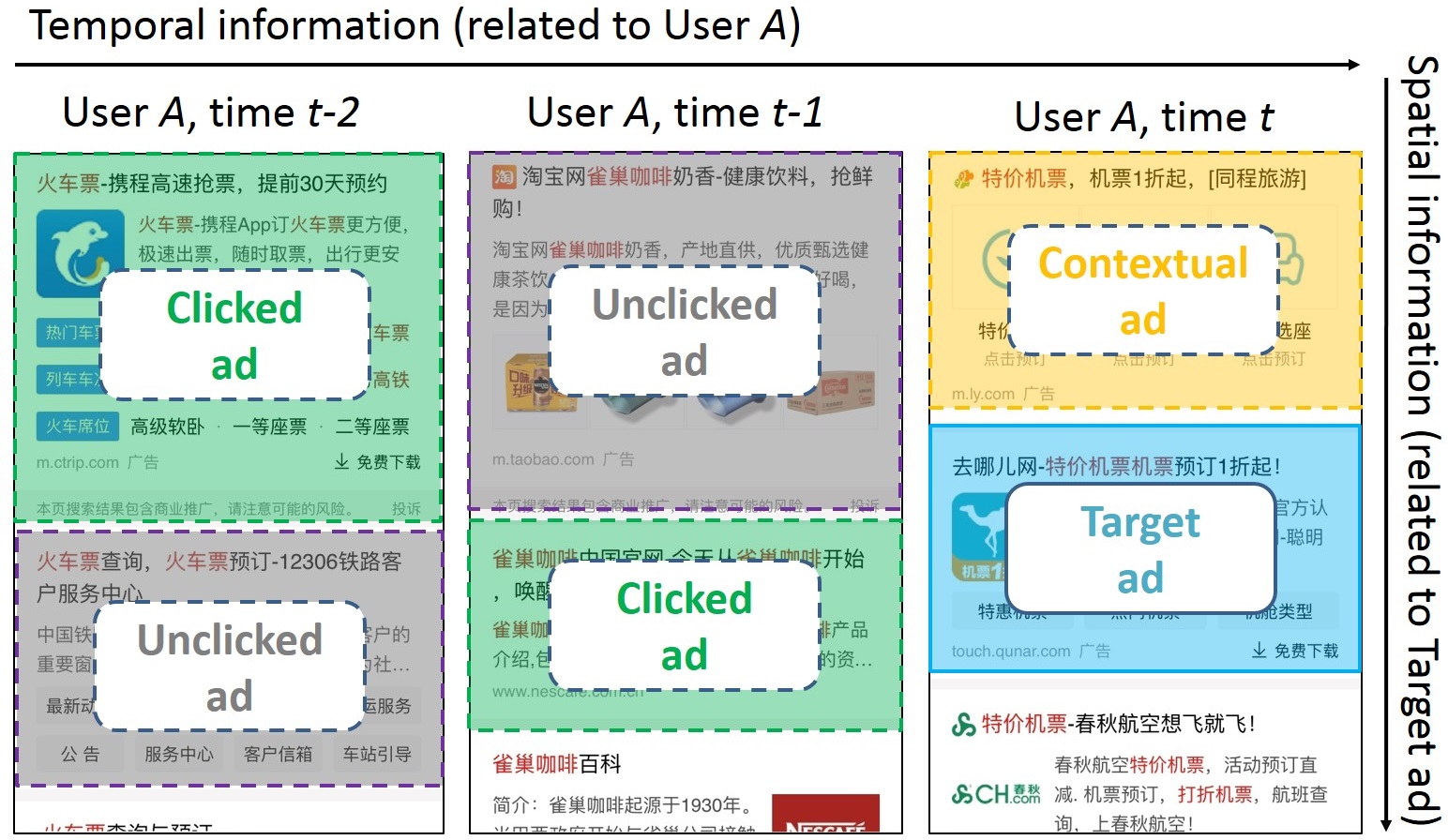}
\vskip -8pt
\caption{Illustration of different types of auxiliary ads for improving the CTR prediction. We consider 1) contextual ads shown above the target ad, 2) historically clicked ads of the user, and 3) historically unclicked ads of the user.}
\vskip -5pt
\label{illus}
\end{figure}

This line of research considers each ad independently, but ignores other ads that may impact the CTR of the target ad.
In this paper, we explore auxiliary ads beyond the target ad for improving the CTR prediction (cf. Figure \ref{illus}). In particular, we explore auxiliary ads from two viewpoints. One is from the spatial domain: we consider the \emph{contextual} ads shown above the target ad on the same page\footnote{Contextual ads are available to advertising systems when ads are ranked sequentially. Please refer to \S\ref{sec_deploy} for more detail.}. The intuition is that ads shown together may compete for a user's attention. 
The second viewpoint is from the temporal domain: we consider the historically \emph{clicked} and \emph{unclicked} ads\footnote{Unclicked ads are ads that are shown to a user but not clicked by the user. They are not created by sampling from an ad pool excluding clicked ads.} of the user. The intuition is that clicked ads reflect a user's preferences and unclicked ads may indicate what a user dislikes to certain extent.

In order to effectively utilize these auxiliary data, we must address the following issues: 1) As the numbers of each type of auxiliary ads may vary, the model must be able to accommodate all possible cases. For example, there are 1 contextual ad, 2 clicked ads and 4 unclicked ads with target ad $a_1$ and there are 0 contextual ad, 3 clicked ads and 2 unclicked ads with target ad $a_2$. 2) As auxiliary ads may not be necessarily related to the target ad, the model should be able to distill useful information and suppress noise in auxiliary data. For example, the clicked ads are about coffee, clothing and car. Then which of these ads are more useful in predicting the target CTR if the target ad is about coffee? 3) The degree of influence of each type of auxiliary ad could be different and the model should be able to differentiate their contributions. For example, the importance of contextual ads and that of clicked ads should be treated differently. 4) The model should be able to fuse all the available information.
To address these issues, we propose three variants of Deep Spatio-Temporal neural Networks (DSTNs) for CTR prediction. These variants include a pooling model, a self-attention model and an interactive attention model, with enhanced ability. The interactive attention model fully addresses the aforementioned issues. Both offline and online experimental results demonstrate the effectiveness of DSTNs for CTR prediction.

\begin{table}[!t]
\caption{Each row is an instance for CTR prediction. The first column is the label, where ``1'' denotes the user clicked the ad and ``0'' otherwise. Each of the other columns is a field. Instantiation of a field is a feature.}
\vskip -8pt
\label{tab_ft}
\centering
\begin{tabular}{|c|c|c|c|c|}
\hline
\textbf{Label} & \textbf{User ID} & \textbf{User Age} & \textbf{Ad Title} \\
\hline
1 & 2135147 & 24 & Beijing flower delivery \\
\hline
0 & 3467291 & 31 & Nike shoes, sporting shoes \\
\hline
0 & 1739086 & 45 & Female clothing and jeans \\
\hline
\end{tabular}
\vskip -8pt
\end{table}

The main contributions of this work are summarized as follows:
\begin{enumerate}
\item We explore three types of auxiliary data for improving the CTR prediction of the target ad. These auxiliary data include contextual ads shown above the target ad on the same page and historically clicked and unclicked ads of the user who views the target ad.
\item We propose DSTNs that effectively fuse these auxiliary data to predict the target CTR. The model is able to learn the interactions between auxiliary data and the target ad, and to emphasize more important hidden information. We make the implementation code publicly available\footnote{https://github.com/oywtece/dstn}.
\item We conduct extensive offline experiments on three large-scale datasets from real advertisement systems to test the performance of DSTNs and several state-of-the-art methods. We also conduct case studies to provide further insights behind the model.
\item We have deployed the best-performing DSTN in Shenma Search, which is the second largest search engine in China. We have also conducted online A/B test to evaluate its performance in real-world CTR prediction tasks.
\end{enumerate}

\section{DSTN Models}
In this section, we first introduce the CTR prediction problem and then present three variants of DSTN models.

\begin{figure}[!t]
\centering
\includegraphics[width=0.36\textwidth]{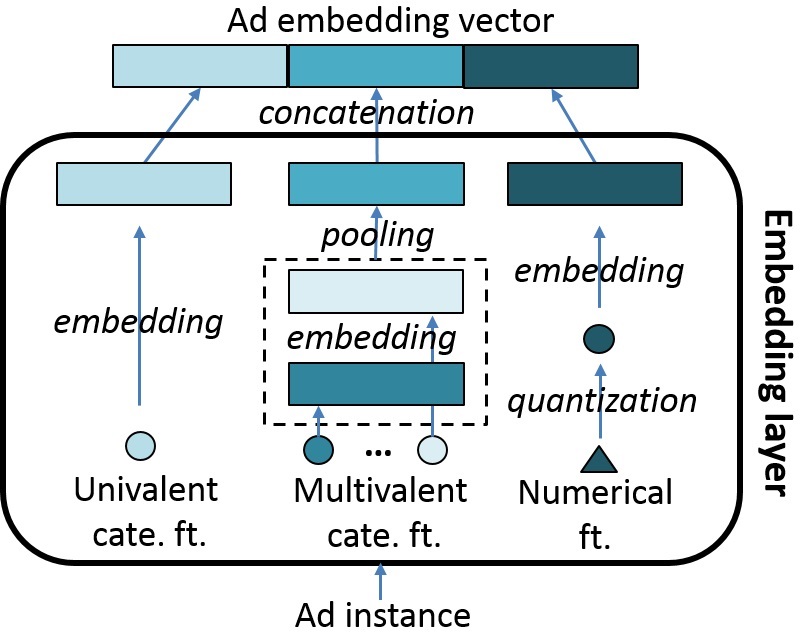}
\vskip -8pt
\caption{Illustration of the embedding process. The representation $\mathbf{x}$ of an ad is the concatenation of all the embedding vectors, each for a field. (cate. - categorical, ft. - feature)}
\vskip -12pt
\label{embed}
\end{figure}

\subsection{Overview}
The task of CTR prediction in online advertising is to build a prediction model to estimate the probability of a user clicking on a specific ad.
Each ad instance can be described by multiple \emph{fields} such as user information (``User ID'', ``City'', ``Age'', etc.) and ad information (``Creative ID'', ``Campaign ID'', ``Title'', etc.). The instantiation of a field is a \emph{feature}. For example, the ``User ID'' field may contain features such as ``2135147'' and ``3467291''. Table \ref{tab_ft} shows some examples.

\begin{figure*}[!t]
\centering
\includegraphics[width=\textwidth]{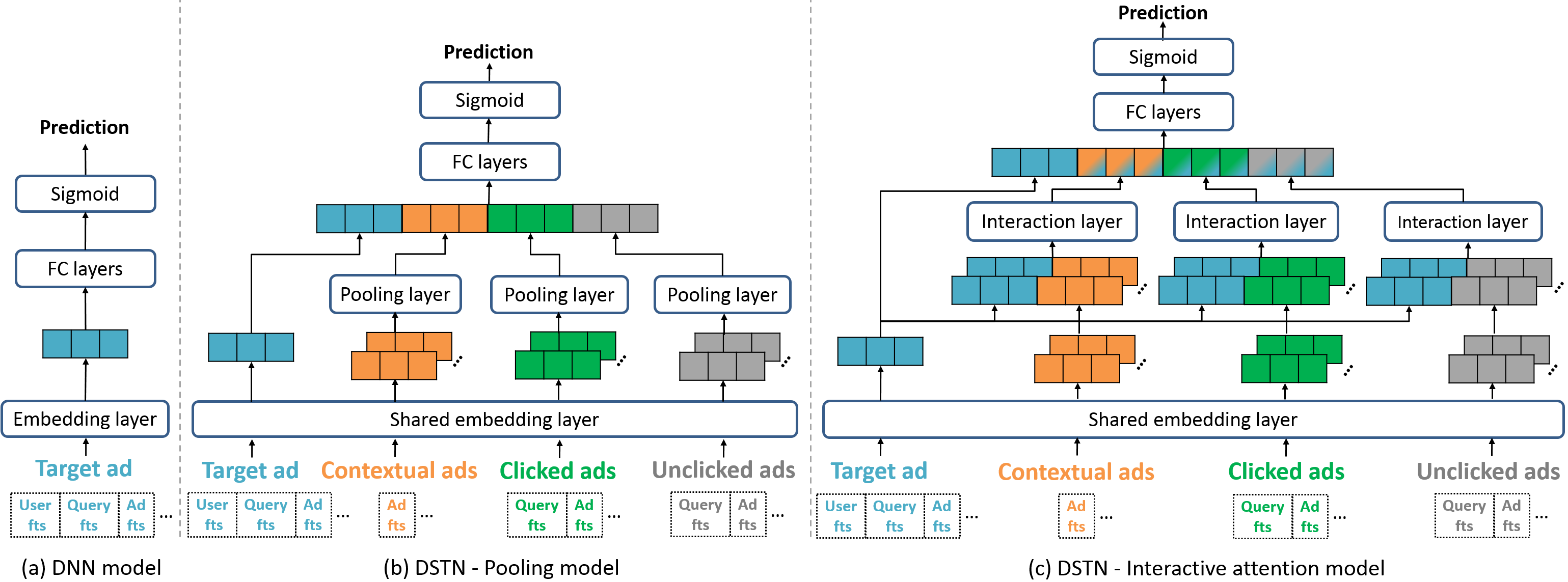}
\vskip -8pt
\caption{Illustration of model architectures (fts - features). (a) DNN model, which considers only the target ad. (b) DSTN - Pooling model, which further considers auxiliary ads. The aggregation of each type of auxiliary ads is by sum pooling. (c) DSTN - Interactive attention model, which introduces explicit interaction between the auxiliary ads and the target ad.}
\vskip -5pt
\label{fig_model}
\end{figure*}

Classical CTR prediction models such as FM \cite{rendle2010factorization}, DNN \cite{zhang2016deep} and Wide\&Deep \cite{cheng2016wide} mainly consider the target ad (illustrated in Figure \ref{fig_model}(a)). The focus is on how to represent the ad instance in terms of informative features and on how to learn feature interactions.
Differently, in this paper, we explore auxiliary data for improving the CTR prediction. We must address the following issues: 1) how to accommodate all different cases with varying numbers of auxiliary ads; 2) how to distill useful information and suppress noise in auxiliary ads; 3) how to differentiate the contributions of each type of auxiliary ads; 4) how to fuse all the available information.

\subsection{Embedding}
Before we introduce any model structure, we first present the embedding process (layer) that is common in all the models below.
The embedding process is to first map each feature into an embedding vector and then represent each ad instance as the concatenation of corresponding feature embedding vectors.

Denote the number of unique features as $N$. We create an embedding matrix $\mathbf{E} \in \mathbb{R}^{N \times K}$ where each row represents a $K$-dim embedding vector for a feature. Given a feature index such as $i$, its embedding is then the $i$-th row of the matrix $\mathbf{E}$. The embedding matrix $\mathbf{E}$ is a variable to be learned during model training.

We segregate features into the following three groups, which are processed differently according to Figure \ref{embed}.

\textbf{1) Univalent categorical features.} This type of feature contributes only a single value and ``User ID'' is an example (cf. Table \ref{tab_ft}). If we use the one-hot feature representation, the resulting feature vector is very sparse because the number of unique user IDs can be on the order of $10^8$. We thus map sparse, high-dimensional categorical features to dense, low-dimensional embedding vectors suitable for neural networks. These embeddings also carry richer information than one-hot representations \cite{mikolov2013distributed}.

\textbf{2) Multivalent categorical features.} This type of feature contributes a set of values and the bag of bi-grams in the ``Ad Title'' is an example (cf. Table \ref{tab_ft}). To illustrate, the bi-grams of title ``ABCD'' are ``AB'', ``BC'' and ``CD''.
As the set cardinality may vary, we first map each value in the set to an embedding vector and then perform sum pooling to generate an aggregated vector of fixed length.

\textbf{3) Numerical features.} ``User Age'' is an example of numerical features (cf. Table \ref{tab_ft}). Each numerical feature is first quantized into discrete buckets, and is then represented by the bucket ID. Each bucket ID is mapped to an embedding vector.

The representation $\mathbf{x}$ of an ad instance after the embedding process is the concatenation of all the embedding vectors, each for a field (cf. Figure \ref{embed}).
Note that the fields used for each type of ads may be different. 
For example, the fields for the target ad contain user information such as ``User ID'' and ``Age'', while the fields for the clicked ads will not contain such information because these ads are for the same user and duplicated information in unnecessary.

After embedding, we obtain one embedding vector $\mathbf{x}_t \in \mathbb{R}^{D_t}$ for the target ad, $n_c$ embedding vectors $\{\mathbf{x}_{ci} \in \mathbb{R}^{D_c} \}_{i=1}^{n_c}$ for the corresponding contextual ads, $n_l$ embedding vectors $\{\mathbf{x}_{lj} \in \mathbb{R}^{D_l}\}_{j=1}^{n_l}$ for clicked ads, and $n_u$ embedding vectors $\{\mathbf{x}_{uq} \in \mathbb{R}^{D_u}\}_{q=1}^{n_u}$ for unclicked ads. $D_\ast$ ($\ast \in \{t, c, l, u\}$) is the vector dimension.

\subsection{DSTN - Pooling Model} \label{sec_base}
Since the numbers $n_c$, $n_l$ and $n_u$ of auxiliary ads may vary for different target ads, it creates a problem for the deep neural network. Therefore, the first issue we need to solve is to process each type of variable-length auxiliary instances into a fixed-length vector.

In the DSTN - Pooling model, we use sum pooling to achieve this goal. The model architecture is shown in Figure \ref{fig_model}(b). The aggregated representation vectors $\mathbf{x}_c$ of $n_c$ contextual ads, $\mathbf{x}_l$ of $n_l$ clicked ads and $\mathbf{x}_u$ of $n_u$ unclicked ads are given by
\[
\mathbf{x}_c = \sum_{i=1}^{n_c} \mathbf{x}_{ci}, \ \mathbf{x}_l = \sum_{j=1}^{n_l} \mathbf{x}_{lj}, \ \mathbf{x}_u = \sum_{q=1}^{n_u} \mathbf{x}_{uq}.
\]
If one type of auxiliary ad is completely missing (e.g., no contextual ads at all), we use an all 0 vector as its aggregated representation.

Now we have the target representation $\mathbf{x}_t$ and the aggregated representations $\mathbf{x}_c$, $\mathbf{x}_l$ and $\mathbf{x}_u$ of auxiliary ads, the next issue is to fuse the information contained in these representations. In particular, we generate the fused representation $\mathbf{v} \in \mathbb{R}^{D_v}$ as
\begin{equation} \label{v}
\mathbf{v} = \mathbf{W}_t \mathbf{x}_t + \mathbf{W}_c \mathbf{x}_c + \mathbf{W}_l \mathbf{x}_l + \mathbf{W}_u \mathbf{x}_u + \mathbf{b},
\end{equation}
where $\mathbf{W}_t \in \mathbb{R}^{D_v \times D_t}$, $\mathbf{W}_c \in \mathbb{R}^{D_v \times D_c}$, $\mathbf{W}_l \in R^{D_v \times D_l}$ and $\mathbf{W}_u \in \mathbb{R}^{D_v \times D_u}$ are weight matrices that transform different representations into the same semantic space; $\mathbf{b} \in \mathbb{R}^{D_v}$ is a bias parameter.

As can be seen, we actually use different weights to fuse the input from different types of data. This property is desired. It is because the degree of influence by different types of auxiliary data to the target ad may be different and we do distinguish such differences. Moreover, the fused representation $\mathbf{v}$ has a property that it is not impacted if one or more auxiliary ads are completely missing. For example, if there is no contextual ads at all, we then set $\mathbf{x}_c = \mathbf{0}$. As a result, we have $\mathbf{W}_c \mathbf{x}_c = \mathbf{0}$ and thus $\mathbf{v}$ is not impacted.

If we concatenate the representations as $\mathbf{m} = [\mathbf{x}_t, \mathbf{x}_c, \mathbf{x}_l, \mathbf{x}_u]$, we can rewrite Eq. (\ref{v}) as $\mathbf{v} = \mathbf{W} \mathbf{m} + \mathbf{b}$, where $\mathbf{W} \in \mathbb{R}^{D_v \times (D_t + D_c + D_l + D_u)}$ is the concatenation of all the weight matrices.
This much simplifies the model. Therefore, the final design is to first concatenate respective representations to obtain an intermediate representation $\mathbf{m}$ and then let $\mathbf{m}$ go through several fully connected layers with the ReLU activation function (defined as $\mathrm{ReLU}(x) = \max(0, x)$), in order to exploit \emph{high-order feature interaction} as well as \emph{nonlinear transformation}.
Nair and Hinton \cite{nair2010rectified} show that ReLU has significant benefits over sigmoid and tanh activation functions in terms of the convergence rate and the quality of obtained results.

Formally, the fully connected layers are defined as follows:
\begin{align}
\mathbf{z}_1 & = \mathrm{ReLU}(\mathbf{W} \mathbf{m} + \mathbf{b}), \
\mathbf{z}_2 = \mathrm{ReLU}(\mathbf{W}_2 \mathbf{z}_1 + \mathbf{b}_2), \ \cdots \nonumber \\
\mathbf{z}_L & = \mathrm{ReLU}(\mathbf{W}_L \mathbf{z}_{L-1} + \mathbf{b}_L),  \nonumber
\end{align}
where $L$ denotes the number of hidden layers; $\mathbf{W}_l$ and $\mathbf{b}_l$ denote the weight matrix and bias vector (to be learned) for the $l$th layer.

Finally, the output vector $\mathbf{z}_L$ goes through a sigmoid function to generate the predicted CTR of the target ad as
\[
\hat{y} = \frac{1}{1+\exp[- (\mathbf{w}^T \mathbf{z}_L + b)]},
\]
where $\mathbf{w}$ and $b$ are the weight and bias parameters to be learned.
To avoid model overfitting, we apply dropout \cite{srivastava2014dropout} after each fully connected layer. Dropout prevents feature co-adaptation by setting to zero a portion of hidden units during parameter learning \cite{goodfellow2013maxout}.

All the model parameters are learned by minimizing the average logistic loss on a training set as
\begin{equation} \label{loss}
\mathrm{loss} = - \frac{1}{|\mathbb{Y}|}\sum_{y \in \mathbb{Y}} [y \log \hat{y} + (1 - y) \log (1 - \hat{y})],
\end{equation}
where $y \in\{0,1\}$ is the true label of the target ad corresponding to the estimated CTR $\hat{y}$ and $\mathbb{Y}$ is the collection of labels.

\subsubsection{Analysis.}
It is observed that when different target ads are shown at a given position for a given user, only $\mathbf{x}_t$ varies while all the auxiliary representations $\mathbf{x}_c$, $\mathbf{x}_l$ and $\mathbf{x}_u$ keep unchanged. It means that the auxiliary representations only serve as static base information.
Moreover, as $\mathbf{x}_c$, $\mathbf{x}_l$ and $\mathbf{x}_u$ are generated by sum pooling, useful information could be easily buried in noise. For example, if the target ad is about coffee but most of the clicked ads are about clothing, then these clicked ads contribute little information to the target ad but the result of sum pooling is clearly dominated by these ads.

\subsection{DSTN - Self-Attention Model} \label{sec_att}
Given the above limitations, we consider the attention mechanism \cite{bahdanau2014neural} that is firstly introduced in the encoder-decoder framework for the machine translation task. It allows a model to automatically search for parts of a source sentence that are relevant to predicting a target word.
In our DSTN - Self-attention model, the attention is applied over the instances of a particular type of auxiliary data to emphasize more important information. Take contextual ads as an example. The aggregated representation $\check{\mathbf{x}}_c$ is modeled as
\begin{equation} \label{att}
\check{\mathbf{x}}_c = \sum_{i=1}^{n_c} \alpha_{ci}(\{\mathbf{x}_{ci}\}_i) \mathbf{x}_{ci},
\end{equation}
where
\[
\alpha_{ci} = \frac{\exp(\beta_{ci})}{\sum_{i'=1}^{n_c} \exp(\beta_{ci'})}, \beta_{ci} = f(\mathbf{x}_{ci}).
\]
$f(\cdot)$ is a function that transforms the vector representation $\mathbf{x}_{ci}$ to a scalar weight $\beta_{ci}$. A possible instantiation of the $f(\cdot)$ function could be a Multilayer Perceptron (MLP) \cite{bahdanau2014neural}.

\subsubsection{Analysis.}
This self-attention mechanism has the advantage that useful information can be emphasized and noise can be suppressed because it weights different auxiliary ads $\mathbf{x}_{ci}$ differently according to (\ref{att}).

Nevertheless, it still has the following limitations: 1) The weight $\beta_{ci}$ is calculated solely based on the auxiliary ad $\mathbf{x}_{ci}$. It does not capture the relationship between this auxiliary ad and the target ad $\mathbf{x}_t$. For example, no matter the target ad $\mathbf{x}_t$ is about coffee or clothing, the importance of auxiliary ads keeps the same.
2) The normalized weight $\alpha_{ci}$ is calculated based on the relative importance with respect to all the $\{\mathbf{x}_{ci}\}_{i=1}^{n_c}$ and $\sum_{i=1}^{n_c} \alpha_{ci} = 1$. As a result, even when all the $\{\mathbf{x}_{ci}\}_{i=1}^{n_c}$ are irrelevant to the target ad $\mathbf{x}_t$, the final importance $\alpha_{ci}$ is still large due to normalization. 3) The absolute number of each type of auxiliary ads matters, but normalization does not capture such an effect either.

\begin{table*}[!th]
\renewcommand{\arraystretch}{1.1}
\caption{Statistics of experimental datasets. (avg - average, ctxt - contextual, pta - per target ad)}
\vskip -8pt
\label{tab_stat}
\centering
\begin{tabular}{|l|c|c|c|c|c|c|c|}
\hline
\textbf{Dataset} & \textbf{\# Target ads} & \textbf{\# Fields} & \textbf{\# Features} & \textbf{Avg \# ctxt ads pta} & \textbf{Avg \# clicked ads pta} & \textbf{Avg \# unclicked ads pta}\\
\hline
Avito & 11,211,794 & 27 & 42,301,586 & 0.9633 & 0.4595 & 4.6739 \\
\hline
Search & 15,007,303 & 20 & 46,529,832 & 0.4456 & 0.7729 & 3.2840 \\
\hline
News Feed & 1,661,588  & 41 & 6,259,571 & N/A & 0.8966 & 2.8853 \\
\hline
\end{tabular}
\end{table*}

\subsection{DSTN - Interactive Attention Model} \label{sec_dstn}
We thus propose the DSTN - Interactive attention model in this section that avoids the above limitations.  It introduces explicit interaction between each type of auxiliary ads and the target ad. The model architecture is shown in Figure \ref{fig_model}(c).

Take contextual ads as an example. We now model the aggregated representation vector as
\begin{equation} \label{interact}
\tilde{\mathbf{x}}_c = \sum_{i=1}^{n_c} \alpha_{ci}(\mathbf{x}_t, \mathbf{x}_{ci}) \mathbf{x}_{ci}.
\end{equation}
Comparing (\ref{interact}) with (\ref{att}), it is observed that $\alpha_{ci}$ is now a function of both the target ad $\mathbf{x}_t$ and the auxiliary ad $\mathbf{x}_{ci}$. In this way, $\alpha_{ci}$ dynamically adjusts the importance of $\mathbf{x}_{ci}$ with respect to $\mathbf{x}_t$.
Moreover, $\alpha_{ci}$ does not depend on other $\{\mathbf{x}_{ci'}\}_{i'\neq i}$. If none of auxiliary ads is informative, then all the $\alpha_{ci}$ should be small. We obtain $\tilde{\mathbf{x}}_l$ and $\tilde{\mathbf{x}}_u$ for clicked ads and unclicked ads similarly.

We learn the weight $\alpha_{ci}$ through an MLP with one hidden layer and the ReLU activation function as
\begin{equation}
\alpha_{ci}(\mathbf{x}_t, \mathbf{x}_{ci}) = \exp (\mathbf{h}^T \mathrm{ReLU}(\mathbf{W}_{tc} [\mathbf{x}_t, \mathbf{x}_{ci}]+ \mathbf{b}_{tc1} ) + b_{tc2}),
\end{equation}
where $\mathbf{h}$, $\mathbf{W}_{tc}$, $\mathbf{b}_{tc1}$ and $b_{tc2}$ are model parameters.

\subsubsection{Analysis.}
In this model, the fused representation $\mathbf{v}$ is
\begin{equation} \label{v2}
\mathbf{v} = \mathbf{W}_t \mathbf{x}_t + \mathbf{W}_c \tilde{\mathbf{x}}_c + \mathbf{W}_l \tilde{\mathbf{x}}_l + \mathbf{W}_u \tilde{\mathbf{x}}_u + \mathbf{b}.
\end{equation}
Comparing (\ref{v2}) with (\ref{v}), it is observed that the auxiliary representations do not serve as static base information now, but dynamically change when the target ad changes. It is because the weights in (\ref{interact}) depends on the target ad as well. It means that the model adaptively distills more useful information in auxiliary data with respect to the target ad. For example, the clicked ads of a user are about coffee, clothing and car. When the target ad $\mathbf{x}_t$ is about coffee, the clicked ad of coffee should contribute more to $\tilde{\mathbf{x}}_l$; but when the target ad is about car, the clicked ad of car should contribute more to $\tilde{\mathbf{x}}_l$. The model also preserves the property that it uses different weights to fuse the input from different types of auxiliary data. Furthermore, the weight $\alpha_{ci}$ in this model is not compared with other auxiliary ads, avoiding the problems that normalization would cause.

\section{Offline Experiments}
In this section, we conduct experiments on three large-scale datasets to evaluate the performance of the proposed DSTNs as well as several state-of-the-art methods for CTR prediction.

\subsection{Datasets}
The statistics of the datasets are listed in Table \ref{tab_stat}. It is observed that the number of distinct features can be up to 46 million.

1) \textbf{Avito advertising dataset.}
This is the dataset used for the Avito context ad clicks competition\footnote{https://www.kaggle.com/c/avito-context-ad-clicks/data}. Note that the ``context ads'' here are different from the contextual ads that we bring forward in this paper.
In Avito, ``context ads'' refer to ads tailored to the user and the context (such as device and time), in contrast to ``regular ads'' that are ordered by the time when they are added and ``highlighted ads'' that are stuck to the top of a page for some period of time.

This dataset contains a random sample of ad logs from avito.ru, the largest general classified website in Russia.
We use the ad logs from 2015-04-28 to 2015-05-18 for training, those on 2015-05-19 for validation, and those on 2015-05-20 for testing.
The features used include 1) ad features such as ad ID, ad title, ad category and ad parent category, 2) user features such as user ID, IP ID, user agent, user agent OS and user device, and 3) query features such as search query, search location, search category and search parameters.

2) \textbf{Search advertising dataset.}
This dataset contains a random sample of ad impression and click logs from a commercial search advertising system in Alibaba. We use ad logs of 7 consecutive days in June 2018 for training, logs of the next day for validation, and logs of the day after the next day for testing.
The features used include 1) ad features such as ad title, ad ID and industry, 2) user features such as user ID, IP ID and user agent, and 3) query features such as query and search location.

3) \textbf{News feed advertising dataset.}
This dataset contains a random sample of ad impression and click logs from a commercial news feed advertising system in Alibaba.
We use ad logs of 7 consecutive days in July 2018 for training, logs of the next day for validation, and logs of the day after the next day for testing.
The features used include 1) ad features such as ad title, ad ID and industry, 2) user features such as user ID and the number of matched ad topics, and 3) cross-features such as AdType-AdResource.
The auxiliary data in this dataset do not contain contextual ads. This is because only one ad is shown on a page in our news feed advertising system.

\subsection{Methods Compared}
We compare the following methods for CTR prediction.
\begin{enumerate}
\item \textbf{LR}. Logistic Regression \cite{bishop2006pattern}. It is a generalized linear model.
\item \textbf{FM}. Factorization Machine \cite{rendle2010factorization}. It models both first-order feature importance and second-order feature interactions.
\item \textbf{DNN}. Deep Neural Network. Each target ad first goes through an embedding layer and then goes through several fully connected layers. Finally, an output layer predicts the CTR through a sigmoid function.
\item \textbf{Wide\&Deep}. The Wide\&Deep model in \cite{cheng2016wide}. It combines LR (wide part) with DNN (deep part).
\item \textbf{DeepFM}. The DeepFM model in \cite{guo2017deepfm}. It combines FM (wide part) with DNN (deep part), and shares the same input and the embedding vector in its wide part and deep part. DeepFM has been shown to outperform Wide\&Deep, Factorization-machine supported Neural Network (FNN) \cite{zhang2016deep} and Product-based Neural Network (PNN) \cite{qu2016product}.
\item \textbf{CRF}. The Conditional Random Field-based method in \cite{xiong2012relational}. It considers both the features of an ad
    and its similarity to the surrounding ads. The predicted log odds of CTR is given by $\mathbf{w}^T \mathbf{x} - 0.5 \beta s$, where $\mathbf{w}$ and $\beta$ are model parameters, $\mathbf{x}$ is the feature vector of the target ad, and $s$ is the sum of similarity to the surrounding ads. The similarity is manually defined on strings in ad titles and ad descriptions. CRF is somewhat impractical because in commercial advertising systems, one cannot know ads below a target ad in advance. Therefore, we only use contextual ads (i.e., ads above the target ad) as the surrounding ads.
\item \textbf{GRU}. The Gated Recurrent Unit \cite{chung2014empirical}, one of the most advanced Recurrent Neural Networks (RNNs). It has been shown to be able to avoid gradient vanishing and explosion problems and to result in better performance than the vanilla RNN. GRU is based on the RNN model proposed in \cite{zhang2014sequential}. It utilizes the sequence of clicked ads of a user.
\item \textbf{DSTN-P}. The DSTN - Pooling model proposed in Section \ref{sec_base}. It uses sum pooling to aggregate auxiliary data.
\item \textbf{DSTN-S}. The DSTN - Self-attention model proposed in Section \ref{sec_att}. It uses self-attention to aggregate auxiliary data.
\item \textbf{DSTN-I}. The DSTN - Interaction attention model proposed in Section \ref{sec_dstn}. It introduces explicit interaction between auxiliary data and the target ad.
\end{enumerate}

Among these methods, CRF, GRU and DSTNs consider auxiliary ads, while all other methods focus on the target ad. In particular, CRF considers surrounding ads, GRU considers clicked ads, and DSTNs consider contextual ads, clicked ads and unclicked ads.

\subsection{Parameter Settings}
We set the dimension of the embedding vectors for each feature as 10, because the number of distinct features is huge.
We set the number of fully connected layers in DNN, Wide\&Deep, DeepFM, GRU and DSTNs as 2, each with dimensions 512 and 256. The dropout ratio is set to 0.5. The hidden dimension of GRU is set to 128. The $f(\cdot)$ function in DSTN-S is an MLP with one hidden layer, with dimension 128. The dimension of $\mathbf{h}$ in DSTN-I is also set to 128. All the methods are implemented in Tensorflow and optimized by the Adagrad algorithm \cite{duchi2011adaptive}. We set the batch size as 128.

We use a user's historical behavior in recent 3 days. To reduce the memory requirement, we further restrict that $n_c \leq 5$, $n_l \leq 5$ and $n_u \leq 5$. That is, if there are no more than 5 clicked ads, we use them all; otherwise, we use the most recent 5.

\subsection{Evaluation Metrics}
We use the following metrics to evaluate the compared methods.
\begin{enumerate}
\item \textbf{AUC}: the Area Under the ROC Curve over the test set. It is a widely used metric for CTR prediction tasks. It reflects the probability that a model ranks a randomly chosen positive instance higher than a randomly chosen negative instance. The larger the better. A small improvement in the AUC is likely to lead to a significant increase in the online CTR \cite{cheng2016wide}.
\item \textbf{Logloss}: the value of Eq. (\ref{loss}) over the test set. The smaller the better.
\end{enumerate}

\begin{table}[!t]
\setlength{\tabcolsep}{3pt}
\renewcommand{\arraystretch}{1.2}
\caption{Test AUC and Logloss on three datasets.}
\vskip -5pt
\label{tab_auc}
\centering
\begin{tabular}{|l|c|c||c|c||c|c|}
\hline
 & \multicolumn{2}{|c||}{\textbf{Avito}} & \multicolumn{2}{|c||}{\textbf{Search}}  & \multicolumn{2}{|c|}{\textbf{News Feed}} \\
\hline
\textbf{Algorithm} & AUC & Logloss & AUC & Logloss & AUC & Logloss \\
\hline
LR & 0.7556 & 0.05918 & 0.7914 & 0.5372 & 0.6098 & 0.4122 \\
\hline
FM & 0.7802 & 0.06094 & 0.8001  & 0.5208  & 0.6119 & 0.4127 \\
\hline
DNN & 0.7816 & 0.05655 & 0.7982 & 0.5240 & 0.6134 & 0.4121 \\
\hline
{\small Wide\&Deep} & 0.7817 & 0.05595 & 0.7992 & 0.5225 & 0.6146 & 0.4120 \\
\hline
DeepFM & 0.7819 & 0.05611 & 0.8008 & 0.5211 & 0.6164 & 0.4113 \\
\hline
CRF & 0.7722 & 0.05989 & 0.7956 & 0.5291 & N/A & N/A \\
\hline
GRU & 0.7835 & 0.05554 & 0.7988 & 0.5224 & 0.6367 & 0.4072 \\
\hline
DSTN-P & 0.8310 & 0.05612 & 0.8162 & 0.5096 & 0.6635 & 0.4008 \\
\hline
DSTN-S & 0.8382 & 0.05456 & 0.8201 & 0.5067 & 0.6659 & 0.3996 \\
\hline
DSTN-I & \textbf{0.8395} & \textbf{0.05448} & \textbf{0.8219} & \textbf{0.5056} & \textbf{0.6679} & \textbf{0.3993} \\
\hline
\end{tabular}
\vskip -8pt
\end{table}

\subsection{Effectiveness}
Table \ref{tab_auc} lists the AUC and Logloss values of different methods. It is observed that Wide\&Deep achieves higher AUC than LR and DNN. Similarly, DeepFM achieves higher AUC than FM and DNN. These results show that combining a wide component and a deep component can improve the prediction power of individual models.
CRF performs much better than LR, because it can be considered as rectifying the LR prediction by a term which summarizes the similarity to surrounding ads. However, the similarity is manually defined based on raw strings, thus suffering from the semantic gap problem.
GRU performs better than LR, FM, DNN, Wide\&Deep and DeepFM on two datasets, because GRU additionally utilizes clicked ads. The improvement is the most obvious on the News Feed ad dataset. This is because users do not submit a query in news feed ads and historical behaviors are quite informative.

It is also observed that DSTN-P outperforms GRU. The reasons are two-fold. First, consecutive actions in a user's behavior sequence may not be well correlated. For example, a user clicked an ad of toothpaste one month ago and clicked ads of snacks and coffee recently. The next clicked ad may be about toothpaste again rather than food, simply because of the need of the user, rather than the correlation with the preceding clicked ads. Therefore, considering the order of a user's behavior may not necessarily help improve the prediction performance. Second, DSTN-P can additionally utilize the information in contextual ads and unclicked ads.

\begin{figure}[!t]
\centering
\subfigure[Absolute AUC improvement]{\includegraphics[width=0.235\textwidth, trim = 0 5 20 20, clip]{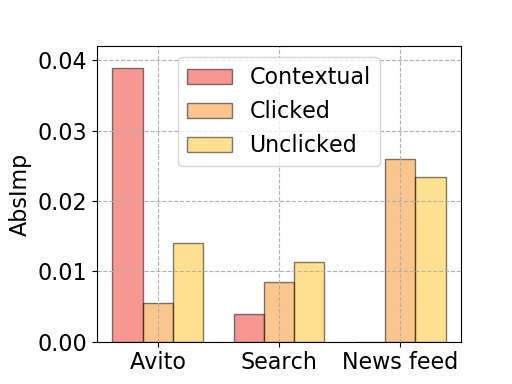}}
\subfigure[Normalized AUC improvement]{\includegraphics[width=0.235\textwidth, trim = 0 5 20 20, clip]{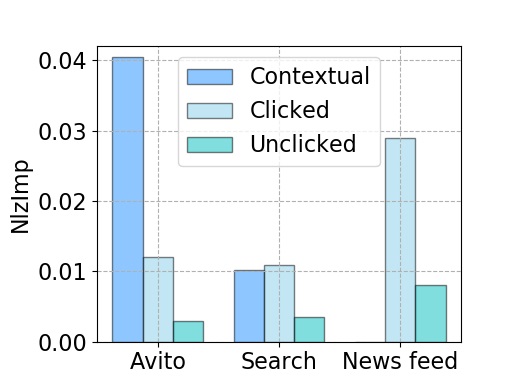}}
\vskip -8pt
\caption{Absolute and normalized AUC improvement: DSTN-I with only one type of auxiliary data, compared with DNN (cf. \S\ref{sec_aux_type}).}
\vskip -8pt
\label{auc_imp}
\end{figure}

\begin{figure}[!t]
\centering
\subfigure[Avito]{\includegraphics[width=0.23\textwidth, trim = 0 0 20 10, clip]{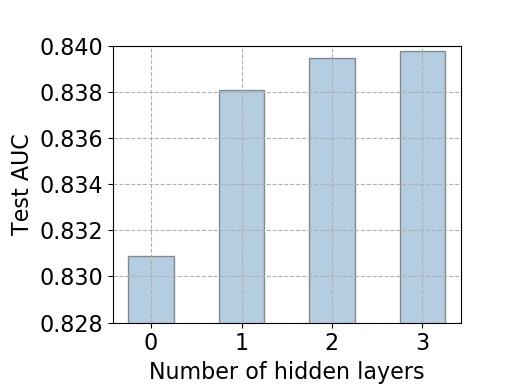}}
\subfigure[News Feed]{\includegraphics[width=0.23\textwidth, trim = 0 0 20 10, clip]{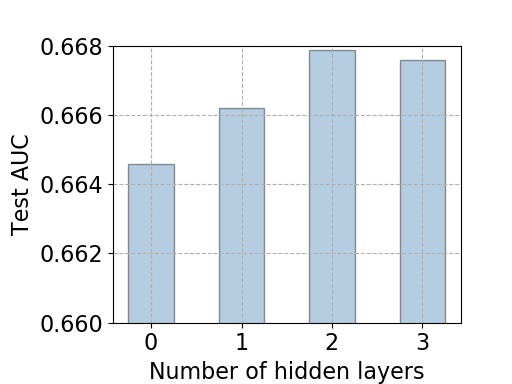}}
\vskip -8pt
\caption{Test AUC vs. the number of fully connected layers of DSTN-I on Avito and News Feed ad datasets.}
\vskip -8pt
\label{n_layer}
\end{figure}

When we compare different variants of DSTNs, it is observed that DSTN-S performs better than DSTN-P, and DSTN-I further outperforms DSTN-S. These results show that self-attention can better emphasize useful information than simple sum pooling. The interactive attention mechanism explicitly introduces the interaction between the target ad and auxiliary ads, and can thus adaptively distill more relevant information than self-attention.

It is also observed that Logloss is not necessarily correlated with AUC. Nevertheless, DSTN-I also results in the smallest Logloss on different datasets, showing its effectiveness.

\subsection{Effect of the Type of Auxiliary Data} \label{sec_aux_type}
In order to examine the effect of different types of auxiliary data, we feed only contextual ads, clicked ads or unclicked ads to DSTN-I.
To quantify the effect, we define and compute the following two metrics: absolute AUC improvement (\emph{AbsImp}) and normalized AUC improvement (\emph{NlzImp}):
\begin{align}
\textrm{\emph{AbsImp}(ctxt)} & = \textrm{AUC(DSTN-I with ctxt ads only)} - \textrm{AUC(DNN)}, \nonumber \\
\textrm{\emph{NlzImp}(ctxt)} &= \frac{\textrm{\emph{AbsImp}(ctxt)}}{\textrm{Avgerage number of ctxt ads per target ad}}, \nonumber
\end{align}
where ctxt is short for contextual.
\emph{AbsImp} considers the overall AUC improvement and \emph{NlzImp} normalizes the effect to each auxiliary ad.
We care about the absolute rather than the relative AUC improvement because in industrial practice the former is more meaningful and indicative.

\begin{figure}[!t]
\centering
\includegraphics[width=0.48\textwidth, trim = 0 0 0 0, clip]{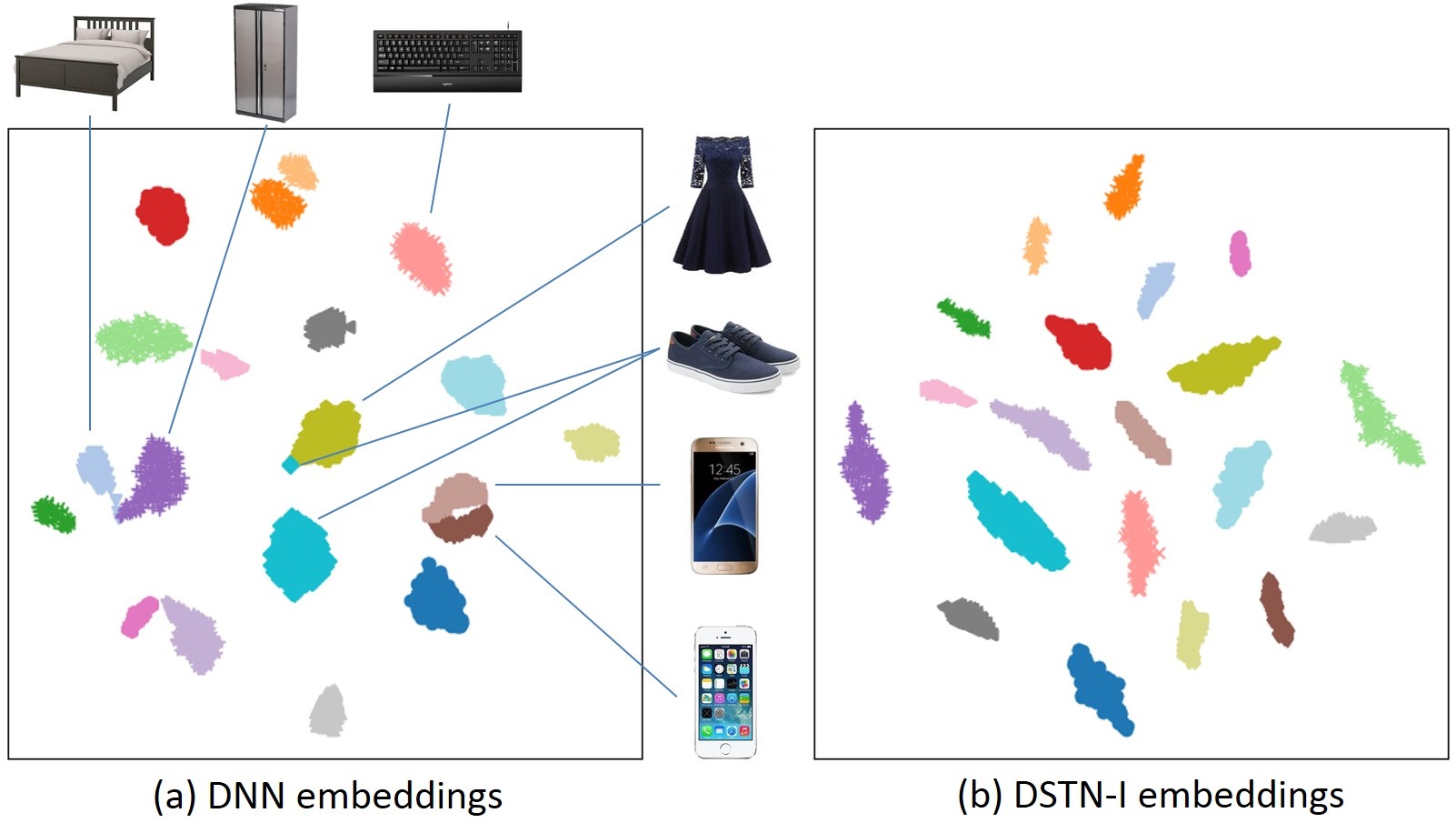}
\vskip -8pt
\caption{Embeddings learned by DNN and by DSTN-I. Each color represents a sub-category of ads (cf. \S\ref{sec_vis_emb}).}
\vskip -5pt
\label{embed_dstn}
\end{figure}

The results are plotted in Figure \ref{auc_imp}. It is observed in Figure \ref{auc_imp}(a) that the overall effect of different types of auxiliary data varies on different datasets. Contextual ads achieve the highest \emph{AbsImp} on the Avito dataset, while unclicked ads achieve the highest \emph{AbsImp} on the Search ad dataset.
From Figure \ref{auc_imp}(b), it is interesting to observe that once normalized, the power of a contextual or clicked ad is much higher than that of an unclicked ad. This complies with intuitions because a contextual ad may distract a user's attention and a clicked ad usually reflects a user's interest. In contrast, an unclicked ad is much noisy. It may indicate that the user is not interested in the ad or the user does not view the ad.

\subsection{Effect of the Network Depth}
Figure \ref{n_layer} plots the test AUC of DSTN-I vs. the number of fully connected layers. The settings are: 1 layer - 256 dimensions; 2 layers - 512 and 256 dimensions; 3 layers - 1024, 512 and 256 dimensions. It is observed that increasing the number of fully connected layers can improve the AUC in the beginning, but the benefit diminishes when more layers are added. Adding more layers may even result in slight performance degradation, possibly due to more model parameters and increased difficulty of training deeper neural networks.

\subsection{Visualization of Ad Embeddings} \label{sec_vis_emb}
Figure \ref{embed_dstn} illustrates the visualization of ad embeddings with t-SNE \cite{maaten2008visualizing} learned by DNN and by DSTN-I. It is based on 20 sub-categories from 5 major categories (electronics, clothing, furniture, computers and personal care). We randomly pick 100 ads in each sub-category. Different colors represent different sub-categories.

It is observed that the embeddings learned by both methods show clear clusters, each representing a group of similar ads. Nevertheless, DNN mixes up a portion of ``iPhone'' vs. ``Samsung phone'', ``beds'' vs. ``cabinets'', and ``dresses'' vs. ``footwear''. In contrast, DSTN-I learns sharper clusters and clearly distinguishes different sub-categories. These results demonstrate that DSTN-I can learn more representative embeddings by the aid of auxiliary ads.

\begin{figure}[!t]
\centering
\includegraphics[width=0.5\textwidth, trim = 0 0 0 0, clip]{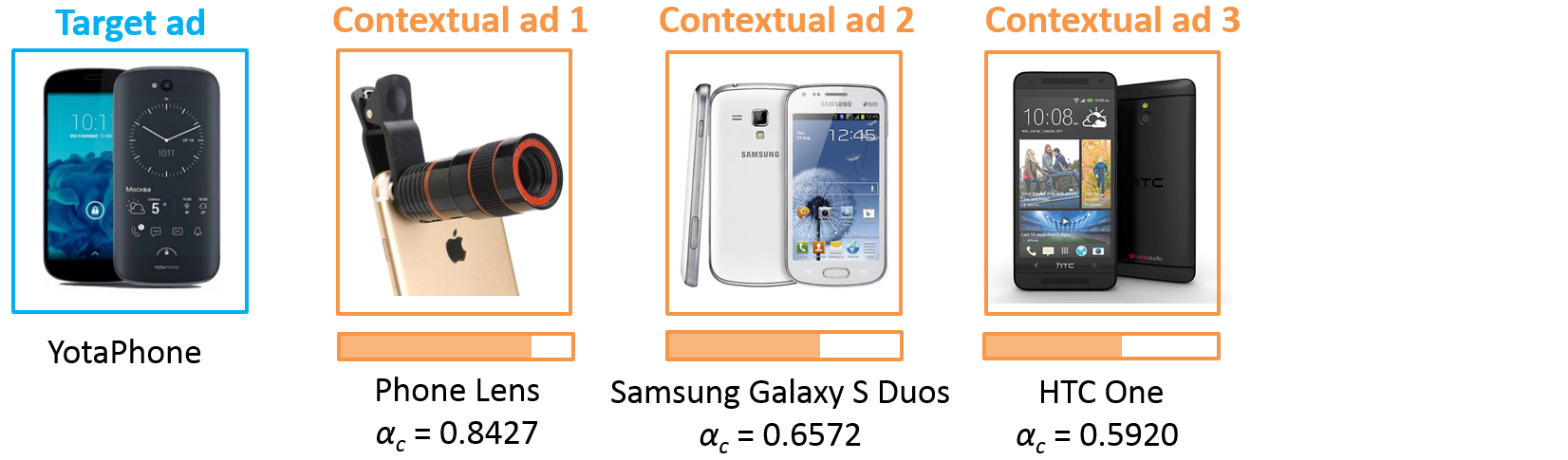}
\vskip -8pt
\caption{Attention weights $\alpha_c$ of contextual ads. The more similar to the target ad, the \emph{smaller} the weight.}
\vskip -6pt
\label{case_ctxt}
\end{figure}

\begin{figure}[!t]
\centering
\includegraphics[width=0.5\textwidth, trim = 0 0 0 0, clip]{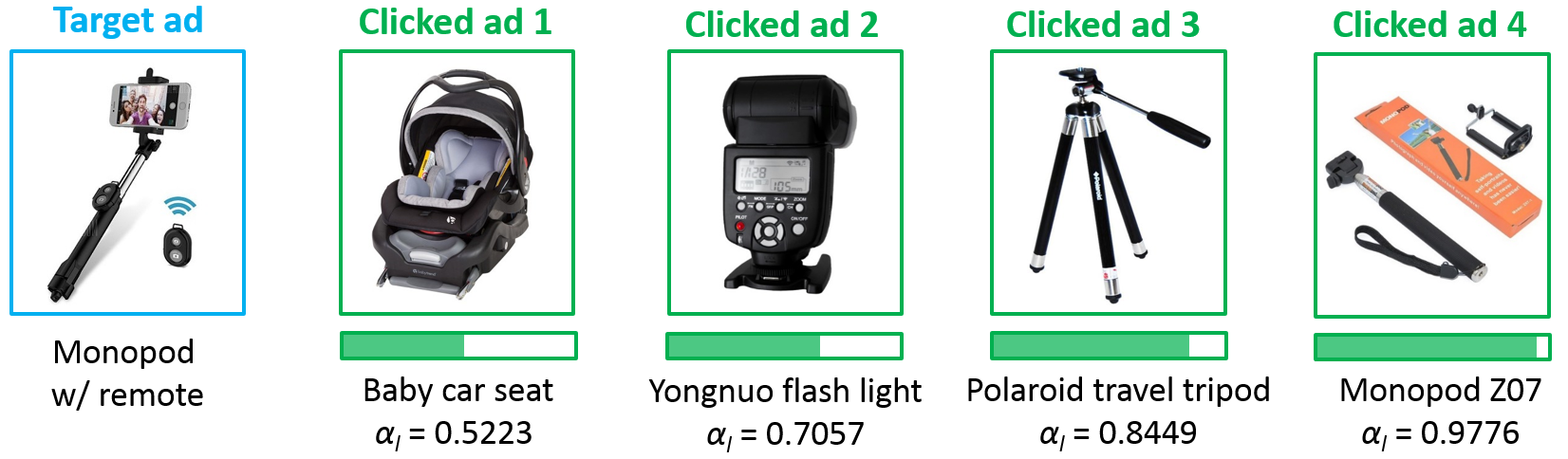}
\vskip -8pt
\caption{Attention weights $\alpha_l$ of clicked ads. The more similar to the target ad, the \emph{larger} the weight.}
\vskip -6pt
\label{case_clk}
\end{figure}

\begin{figure}[!t]
\centering
\includegraphics[width=0.5\textwidth, trim = 0 0 0 0, clip]{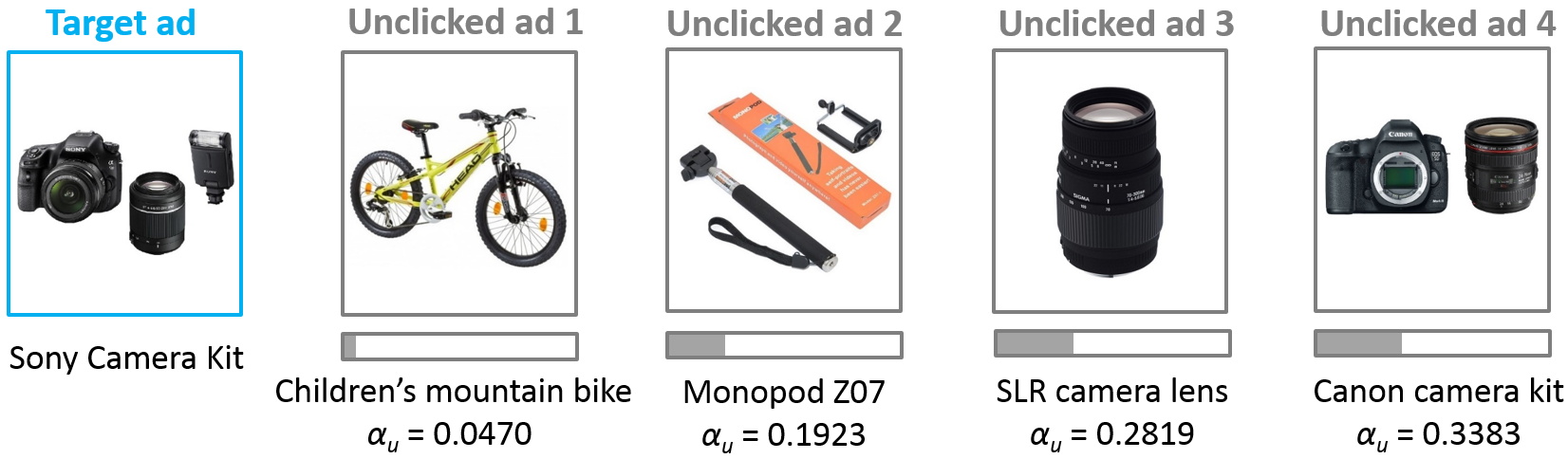}
\vskip -8pt
\caption{Attention weights $\alpha_u$ of unclicked ads. The more similar to the target ad, the \emph{larger} the weight.}
\vskip -6pt
\label{case_unclk}
\end{figure}

\subsection{Visualization of Attention Weights}
In this section, we examine the attention weights of auxiliary ads in DSTN-I through several case studies on the Avito dataset. We examine each type of auxiliary ads separately because it is hard to find a case containing sufficient ads of all types. For ease of illustration, we sort the auxiliary ads by their semantic similarity to the target ad. The leftmost is the most dissimilar and the rightmost is the most similar.

\textbf{Contextual ads.} In Figure \ref{case_ctxt}, the target ad is about a YotaPhone. Three contextual ads are shown, which are about a phone lens, a Samsung phone and a HTC phone. It is observed that the weights of the two phone ads do not differ much (around 0.6), but the weight of the lens ad (most dissimilar) is much higher (over 0.8). Such an observation complies with the analysis in \cite{xiong2012relational}, where the authors find that the more similar the surrounding ads are to an ad, the lower the CTR of the ad is. This is because similar ads can distract a user's attention since all these ads offer similar products or services. In contrast, a dissimilar ad can help make the target ad more notable and is thus assigned a larger weight by DSTN-I.

\textbf{Clicked ads.} In Figure \ref{case_clk}, the target ad is about monopod and remote for self-portrait photograph. The first clicked ad is about a baby car seat, which is clearly not relevant to the target ad. Its weight is 0.5223. The second clicked ad is about a flash light, which is an accessory of digital cameras for photography; its weight is much higher (0.7057). The third click ad is about a tripod which is more similar to a monopod, and thus the weight is even higher (0.8449). Finally, the fourth clicked ad is also about a monopod and its weight is the highest (0.9776). These observations show that the more similar a clicked ad is to the target ad, the higher its weight during aggregation. This is because if a user has clicked a similar ad, it is likely that the user will click the target ad as well.

\begin{figure}[!t]
\centering
\includegraphics[width=0.45\textwidth, trim = 0 0 0 0, clip]{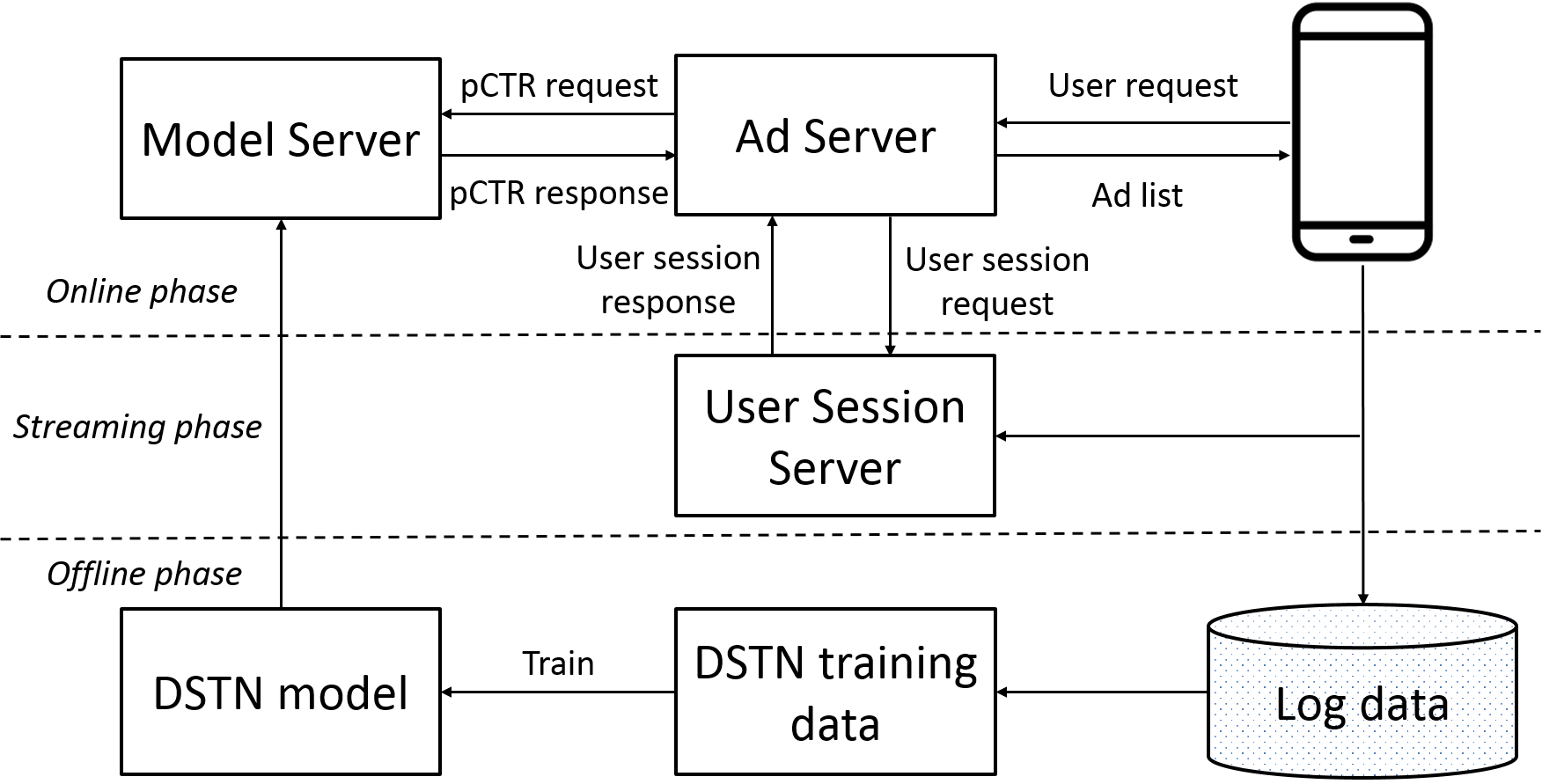}
\vskip -8pt
\caption{Architecture of the online advertising system. The deployed DSTN model is DSTN-I.}
\vskip -11pt
\label{server}
\end{figure}

\textbf{Unclicked ads.} In Figure \ref{case_unclk}, the target ad is about a Sony camera kit. The four unclicked ads are about a bike, a monopod, a camera lens and a camera kit respectively. These ads have increased similarity to the target ad and the corresponding weights also increase. These observations show that the more similar an unclicked ad is to the target ad, the higher its weight during aggregation. It is because if a user does not click a similar ad in the past, it is likely that the user will also not click the target ad as well.

Comparing Figure \ref{case_unclk} with Figure \ref{case_clk}, it is interesting to observe that the average weight of unclicked ads are much smaller than that of clicked ads, even when an unclicked ad is quite similar to the target ad. This is because clicked ads reflect possible user preferences while unclicked ads are much more ambiguous.

\section{System Deployment and Evaluation}
In this section, we introduce the deployment of our advertising system and present the online evaluation results.

\subsection{Deployment} \label{sec_deploy}
\begin{figure}[!t]
\centering
\includegraphics[width=0.48\textwidth, trim = 0 0 0 0, clip]{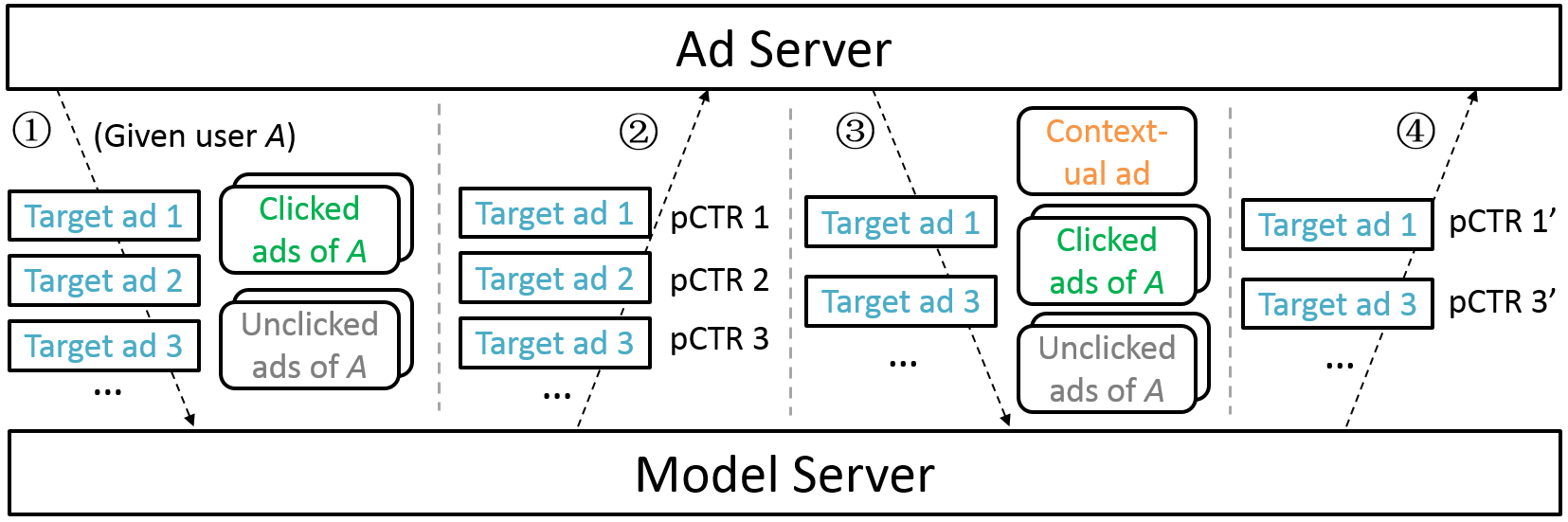}
\vskip -8pt
\caption{Illustration of how the Ad server requests pCTRs from the Model server. Please refer to \S\ref{sec_deploy} for more detail.}
\vskip -11pt
\label{process}
\end{figure}

We deployed DSTN-I in Shenma Search, the second largest search engine in China. We call the model DSTN hereafter for simplicity. Figure \ref{server} depicts the architecture of the system, which contains an offline phase, a streaming phase and an online phase.
\begin{itemize}
\item \textbf{Offline phase}: Online user behaviors (ad impression / click) are continuously logged into the user log database. The system exacts training data from the log database and trains the DSTN model. The offline training is first performed in a batch manner, and the resulting model is then updated incrementally and periodically using recent log data.
\item \textbf{Streaming phase}: Online user behaviors are also sent to the User Session Server (with a delay of no more than 10 seconds), where a hashmap for each user is maintained and updated. To reduce the memory requirement and the online computation load, the hashmap records at most 5 clicked ads and 5 unclicked ads in the recent 3 days for each user.
\item \textbf{Online phase}: Once a user request is sent, the Ad Server first retrieves the user history data from the User Session Server. The Ad Server then requests the Model Server for pCTR (predicted CTR) of a set of candidate ads. This is done in several steps as shown in Figure \ref{process}.
\end{itemize}

The steps are: \textcircled{1} The Ad Server sends the set of candidate target ads, along with the clicked and unclicked ads of the given user to the Model Server. There is no contextual ad now. \textcircled{2} The Model Server returns the pCTRs. \textcircled{3} The Ad Server picks out the target ad with the highest score based on certain strategy (which depends on the pCTR). Assume this ad is the Target ad 2. Then it becomes the contextual ad for other target ads. The Ad Server then sends the remaining target ads, along with the contextual ad, the clicked and unclicked ads of the user to the Model Server. \textcircled{4} The Model Server returns the pCTRs for the remaining candidate ads.

Theoretically, Steps \textcircled{3} and \textcircled{4} could be performed several times to pick out ads one by one and to update contextual ads sequentially. However, there exists a tradeoff between prediction accuracy and service delay. Therefore, in our current implementation, we only perform Steps \textcircled{3} and \textcircled{4} once. After Step \textcircled{4}, the Ad Server picks out 2-3 remaining candidate ads with highest scores and sends the final ad list to the user. Shenma Search now holds 3-4 ad slots per mobile page.

\subsection{Online A/B Test}
We conducted online experiments in an A/B test framework over two weeks in Jan. 2019. The benchmark model is Wide\&Deep, which is our last online serving model. Our online evaluation metric is the real CTR, which is defined as the number of clicks over the number of ad impressions. A larger online CTR indicates the enhanced effectiveness of a CTR prediction model.
We observe that DSTN outperforms Wide\&Deep consistently, resulting in an increase of daily online CTR from 5.13\% to 9.72\%. The average CTR increase is 6.92\%. These results demonstrate the effectiveness of DSTN in practical CTR prediction tasks.

\section{Related Work}
\textbf{CTR prediction.}
Learning the effect of feature interactions seems to be crucial for accurate CTR prediction \cite{rendle2010factorization}.
Generalized linear models, such as Logistic Regression (LR) \cite{richardson2007predicting} and Follow-The-Regularized-Leader (FTRL) \cite{mcmahan2013ad}, have shown decent performance in practice. However, a linear model lacks the ability to learn sophisticated feature interactions \cite{chapelle2015simple}. Factorization Machines (FMs) \cite{rendle2010factorization, blondel2016higher} are proposed to model pairwise feature interactions in terms of the latent vectors corresponding to the involved features. Field-aware FM \cite{juan2016field} and Field-weighted FM \cite{pan2018field} further improve FM by considering the impact of the field that a feature belongs to.

In recent years, Deep Neural Networks (DNNs) have shown powerful ability of automatically learning informative feature representations \cite{lecun2015deep}. DNNs are thus also exploited for CTR prediction and item recommendation in order to automatically learn feature representations and high-order feature interactions \cite{van2013deep,covington2016deep,wang2017deep}. Factorization-machine supported Neural Network (FNN) \cite{zhang2016deep} pre-trains an FM before applying a DNN. Product-based Neural Network (PNN) \cite{qu2016product} introduces a product layer between the embedding layer and the fully connected layer. The Wide\&Deep model \cite{cheng2016wide} combines LR and DNN to capture both low- and high-order feature interactions. Such a structure also improves both the memorization and generalization abilities of the model. DeepFM \cite{guo2017deepfm} models low-order feature interactions like FM and models high-order feature interactions like DNN. Neural Factorization Machine \cite{he2017neural} combines the linearity of FM and the non-linearity of neural networks.

\textbf{Exploiting auxiliary data for CTR prediction.}
Another line of research exploits auxiliary data for improving the CTR prediction performance.
Zhang et al. \cite{zhang2014sequential} consider users' historical behaviors (e.g., what ads she clicked). They use Recurrent Neural Networks (RNNs) to model the dependency on users' sequential behaviors. Tan et al. \cite{tan2016improved} propose improved RNNs for session-based recommendations. One major problem with RNN-based models is that it generates an overall embedding vector of a behavior sequence, which can only preserve very limited information of a user. Long-term dependencies are still hard to be preserved even using the advanced memory cell structures like Long Short-Term Memory (LSTM) \cite{hochreiter1997long} and Gated Recurrent Unit (GRU) \cite{chung2014empirical}. Moreover, both the offline training and the online prediction process of RNNs are time-consuming, due to its recursive structure.

Xiong et al. \cite{xiong2012relational} consider the pairwise relationship between ads shown on the same page and propose a Conditional Random Field (CRF)-based model for CTR prediction. Yin et al. \cite{yin2014exploiting} consider various contextual factors such as ad depth, query diversity and ad interaction for click modeling. One major problem of these models is that the vertex and edge feature functions need to be manually defined based on data analysis and it is difficult to generalize the model to other types of data.

\textbf{Differences.} DSTNs proposed in this paper differ from prior work in that: 1) DSTNs integrate heterogeneous auxiliary data (i.e., contextual, clicked and unclicked ads) in a unified framework, while the RNN-based model \cite{zhang2014sequential} cannot utilize contextual and unclicked ads, and the CRF-based model \cite{xiong2012relational} cannot incorporate clicked and unclicked ads; 2) DSTNs are not based on RNNs and they are much easier to implement and are much faster to train and evaluate online.

\section{Conclusion}
In this paper, we address the problem of CTR prediction in online advertising systems. In contrast to classical CTR prediction models that focus on the target ad, we explore three types of auxiliary data (i.e., contextual, clicked and unclicked ads) and propose DSTNs for improving the CTR prediction. DSTNs are able to distill useful information in auxiliary ads and to fuse heterogeneous data in a unified framework. Offline experimental results on three large-scale datasets demonstrate the effectiveness of DSTNs over several state-of-the-art methods. Case studies show that DSTN-I is able to learn representative ad embeddings and meaningful attention weights. We have deployed DSTN-I in Shenma Search. Online A/B test results show that the online CTR is also improved compared to our last serving model, demonstrating the effectiveness of DSTN-I in real-world CTR prediction tasks.

\bibliographystyle{ACM-Reference-Format}
\bibliography{ref}

\end{document}